# Cross-Lingual Mental Health Ontologies for Indian Languages: Bridging Patient Expression and Clinical Understanding through Explainable AI and Human-in-the-Loop Validation

**Ananth Kandala[1]**  **Ratna Kandala[2]**  **Akshata Kishore Moharir[3]**
**Niva Manchanda[2]**  **Sunaina Singh[4]**

[1]University of Florida  [2]University of Kansas  [3]University of Maryland
[4]IIT Kharagpur

{ananthkandala46, ratnanirupama}@gmail.com, akshatankishore5@gmail.com
nmanchanda@ku.edu, sunainasingh.rathod@gmail.com

## Abstract

Mental health communication in India is linguistically fragmented, culturally diverse, and often underrepresented in clinical NLP. Current health ontologies and mental health resources are dominated by diagnostic frameworks centered on English or Western culture, leaving a gap in representing patient distress expressions in Indian languages. We propose cross-linguistic graphs of patient stress expressions (CL-PDE), a framework for building *cross-lingual mental health ontologies* through graph-based methods that capture culturally embedded expressions of distress, align them across languages, and link them with clinical terminology. Our approach addresses critical gaps in healthcare communication by grounding AI systems in culturally valid representations, allowing more inclusive and patient-centric NLP tools for mental health care in multilingual contexts.

## 1 Introduction

Access to mental health care in India faces systemic barriers beyond infrastructure gaps, with linguistic fragmentation and cultural divergence in symptom expression creating critical bottlenecks in patient-clinician interactions. Although resource scarcity is well documented, the language gap between patients and clinical Natural Language Processing (NLP) systems remains understudied, representing a critical NLP challenge.

Patients describe distress using idioms, metaphors, and culture-bound terms that lack direct English or clinical equivalents. For instance, expressions in Hindi such as *mera mann chintit hai* (I am feeling anxious), *mujhe mansik tanaav mehsoos ho rha hai* (I feel mentally stressed), *mujhe ghabraahat mehsoos ho rhi hai*) (I am anxious), *man ka bhoj* (burden on the mind/heart) carry deep cultural significance but are absent from Western medical taxonomies. Standard NLP tools are trained primarily on the mental health corpora of Western English and do not capture these signals, exacerbating healthcare inequities.

The problem manifests in three critical dimensions:

1. **Low-Resource Language Barriers:** Despite India having one of the largest and fastest growing digital user bases in the world (Statista, 2020), natural language technologies still struggle to serve its population effectively. This gap is striking given the linguistic richness of the region - 22 scheduled languages covering more than 1.17 billion speakers, and 121 languages each having communities larger than 10,000 speakers. In total, 1369 rationalized languages and dialects are spoken across the country (Government of India, 2011). State-of-the-art multilingual systems remain sub-optimal in Indian languages, highlighting the mismatch between technological progress and societal need (Khanuja et al., 2021), including Hindi (Prakash et al., 2024).

2. **Cultural Ontology Mismatch:** Conventional western ontologies (DSM5, ICD11) fail to capture certain culture-specific distress concepts, creating semantic blind spots (Kirmayer et al., 2017; Paniagua, 2018). These frameworks miss nuanced expressions of mental distress that are prevalent in Indian cultural contexts.

3. **Code-Mixing and Dialectal Variation:** Hybrid utterances such as *mujhe stress mehsoos ho raha hai, mujhe anxiety ho rahi hain, tension ho rahi hain, mera mood off hain* challenge monolingual tokenizers, reducing clinical intent detection accuracy and complicating automated assessment tools.

To address these gaps, this paper introduces the Cross-Lingual Graphs of Patient Distress Expressions (CL-PDE), a comprehensive framework for building and utilizing multilingual mental health ontologies while preserving cultural semantics and supporting clinical relevance.

Our contributions include two-fold: (A) A novel graph-based framework for constructing cross-lingual mental health ontologies that preserve cultural semantics and (b) a human-in-the-loop validation methodology that integrates cultural authenticity with clinical expertise.

We argue that **cross-lingual, culturally grounded mental health ontologies** are essential for bridging the language patients use to express distress with the standardized vocabularies on which healthcare systems depend. By developing these resources, we aim to enable more inclusive, patient-centric NLP tools that can strengthen communication between patients and clinicians across linguistic and cultural divides.

This paper is structured as follows: Section 2 reviews prior work. Section 3 introduces the conceptual framework for cross-lingual mental health ontologies. Section 4 outlines the proposed methodology for implementation and evaluation, followed by Section 5, which addresses limitations, and Section 6, which concludes with future directions.

## 2 Prior Work

### 2.1 Clinical NLP and Mental Health

Recent advances in clinical NLP have primarily focused on English-language resources, creating significant barriers for multilingual populations. Transformer-based models have been applied to detect depression from social media posts (Zhang et al., 2022), and early warning systems for mental health conditions have been developed using Reddit data (Yates et al., 2017). More recently, (Atapattu et al., 2022) developed the first emotion-annotated mental health corpus in English, establishing benchmarks for computational approaches to mental health assessment. Despite these contributions, current methods remain grounded in English corpora and Western diagnostic frameworks, limiting their relevance and portability to multilingual and non-Western settings.

The issue of cultural bias in computational mental health has been noted but remains unresolved. (Harrigian et al., 2020) identified cultural bias in mental health detection systems, yet offered no multilingual strategies. Similarly, (Chancellor and De Choudhury, 2020) emphasized the role of cultural context, but their analysis centered on demographic rather than linguistic diversity, leaving the core language gap unaddressed.

(Dissanayake et al., 2020) noted the limited use and development of high-quality clinical reasoning ontologies (CROs) in clinical decision support systems (CDSSs), emphasizing the need for structured knowledge representation in healthcare applications. This gap is particularly pronounced in cross-cultural contexts where standard ontologies fail to capture culturally specific expressions of distress.

### 2.2 Multilingual Health Resources

Efforts to create multilingual health resources have emerged but remain limited in scope and coverage. (Névéol et al., 2018) developed clinical NLP tools for languages beyond English, focusing primarily on European languages with well-established medical terminology databases. (Liu et al., 2021) created multilingual medical knowledge graphs using visual pivoting techniques, but these efforts provided limited coverage of mental health terminology and lacked cultural contextualization.

For Indian languages specifically, progress has been minimal. (Seetha et al., 2007) developed basic health information extraction tools for Hindi, but these systems lack mental health-specific vocabularies and fail to capture the rich cultural expressions of psychological distress prevalent in Indian languages. The scarcity of annotated mental health corpora in Indian languages remains a significant bottleneck for developing effective NLP tools.

A recent Telugu–English code-mixed corpus captures medical dialogue (Dowlagar and Mamidi, 2023), reflecting the multilingual reality of Indian healthcare, but systematic strategies for handling such linguistic complexity in mental health remain unexplored. The absence of culturally grounded corpora continues to block NLP progress in this domain and systematic approaches to handling such linguistic diversity in mental health applications remain largely unexplored, leaving a critical gap in healthcare accessibility.

### 2.3 Cultural Psychiatry and Language

The field of cultural psychiatry has long recognized the fundamental importance of language in mental health expression and diagnosis. (Kleinman, 1991) introduced the seminal concept of "idioms of distress"—culturally specific ways of experiencing and expressing emotional suffering that often lack direct equivalents in Western psychiatric ter-minology. This work established the theoretical foundation for understanding how cultural context shapes mental health communication.

Building on this foundation, (Kohrt and Hruschka, 2010) documented how Nepali expressions of heart-mind distress map poorly onto Western depression constructs, demonstrating the inadequacy of direct translation approaches in cross-cultural mental health assessment. Their ethnographic work revealed that concepts like *man dukheko* (heart-mind pain) encompass spiritual, social, and somatic dimensions that are lost when reduced to Western diagnostic categories.

Recent computational approaches have begun incorporating cultural considerations but remain limited in scope. (Choudhury et al., 2017) explored cross-cultural differences in depression expression on social media, revealing significant variations in how different cultural groups articulate psychological distress online. (Aggarwal et al., 2014) called for integrating cultural concepts into psychiatric assessment and developed frameworks for cultural adaptation of psychological treatments, emphasizing the need for culturally grounded diagnostic tools.

However, systematic frameworks for building culturally grounded computational ontologies that can bridge patient expressions with clinical terminology remain underdeveloped. The translation of cultural psychiatry insights into computational tools capable of supporting clinical practice represents a significant unmet need.

### 2.4 Graph-Based Ontology Alignment

Graph-based methods for ontology alignment have shown considerable promise in medical domains, offering structured approaches to knowledge representation and cross-domain mapping. (Kolyvakis et al., 2018) used graph neural networks for biomedical ontology matching, demonstrating the effectiveness of embedding-based approaches for capturing semantic relationships between medical concepts.

(Liu et al., 2021) developed cross-lingual entity alignment techniques using knowledge graphs, employing visual pivoting methods to establish correspondences between entities across different languages. Their approach showed promise for multilingual knowledge integration but was not specifically designed for healthcare applications. (Trisedya et al., 2019) proposed multilingual knowledge graph completion methods that leverage attribute embeddings for cross-lingual entity alignment, contributing to the technical foundation for multilingual ontology construction.

Despite these advances, existing graph-based approaches have not addressed the unique challenges of culturally sensitive mental health terminology. The incorporation of human validation for cultural authenticity—a critical requirement for healthcare applications—remains absent from current technical solutions. Additionally, the explainability requirements for clinical applications, where practitioners must understand and trust AI-generated interpretations, have not been adequately addressed in existing graph-based ontology alignment work.

The gap between technical capability and clinical applicability in cross-cultural mental health represents a significant opportunity for advancing both computational linguistics and healthcare accessibility.

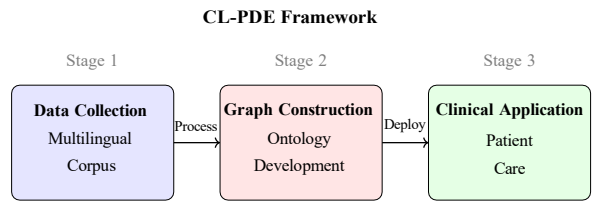

Figure 1: An overview of the Cross-Lingual Graphs of Patient Distress Expressions (CL-PDE) framework

## 3 Proposed Framework

We propose **Cross-Lingual Graphs of Patient Distress Expressions (CL-PDE)**, a comprehensive framework to build and use multilingual mental health ontologies. Figure 1 summarizes the workflow.

### 3.1 Corpus Construction

The foundation of the CL-PDE framework is a corpus of patient narratives collected from various sources, including counseling transcripts, mental health helplines, online forums, and community health worker interactions in multiple Indian languages. Each expression of psychological state, ranging from anxiety and grief to stress and hopelessness, is annotated with linguistic markers and cultural context indicators. Drawing from various sources, the corpus comprehensively captures socioeconomic and regional diversity, ensuring that the ontology does not disproportionately reflect urban or digitally literate populations. Although this corpus captures the diversity of how distress is expressed in languages and contexts, raw narratives alone cannot support clinical or computational use. What is needed is a systematic representation that preserves cultural nuance while enabling alignment with standardized frameworks.

## 3.2 From Narratives to Ontology

To achieve this, we model the data as a heterogeneous graph. Once these expressions are collected, the challenge lies in structuring them so that their cultural richness is preserved while enabling systematic clinical interpretation. To this end, the data are organized as a heterogeneous graph - a natural fit for representing both the diversity of patient expressions and their links to formal mental health ontologies.

In this graph, two kinds of nodes are created: (a) Expression Nodes: which represent culture-bound idioms and metaphors of psychological states. (b) Concept nodes: which represent diagnostic categories drawn from resources such as ICD-11 and DSM-5, including culturally sensitive constructs such as the DSM-5 Cultural Concepts of Stress (Center for Substance Abuse Treatment (US), 2014).

Edges between nodes encode different kinds of relationship: intra-lingual links group related expressions within a single language; cross-lingual links align equivalent expressions across languages; and expression-concept links tie everyday patient language to standardized clinical categories. Each edge is further annotated with metadata (relation type, confidence, provenance) to preserve transparency and allow downstream validation.

This layered representation allows clusters of culturally grounded expressions to co-exist even when no direct clinical equivalent exists, while still providing pathways to analog with standardized psychiatric frameworks. Figure 2 illustrates the multilayered graph structure with example mappings across languages. However, deciding which expressions should be connected to which concepts is not trivial. Direct mappings are often uncertain, context-dependent, or subjective. This motivates our next step: to integrate graph construction with multilingual LLMs and human-in-the-loop validation.

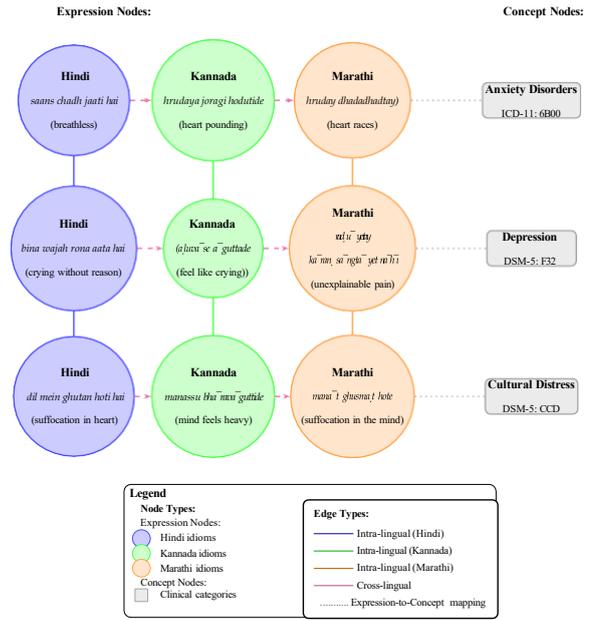

Figure 2: Heterogeneous graph representing culture-bound idioms of mental distress across Hindi, Kannada, and Marathi. The graph contains two node types: (a) Expression nodes (circles) capturing authentic patient narratives, and (b) Concept nodes (rectangles) representing standardized clinical categories from ICD-11 and DSM-5. Three edge types preserve cultural richness: intra-lingual edges (colored solid lines) connect related expressions within each language; cross-lingual edges (red dashed) align equivalent expressions across languages; and expression-to-concept edges (black dotted) link patient language to formal diagnostic frameworks.

## 3.3 Graph–LLM Integration and Human-in-the-Loop Validation

Although the graph structure enables cultural expressions to be systematically represented alongside clinical categories, determining the correct links between nodes is far from trivial. Expressions can be polysemous, context-dependent, and contested even among experts. To address this challenge, our framework combines the generative capacity of multilingual LLM's with structured expert review.

Multilingual LLMs fine-tuned in health-related corpora are first used to suggest candidate edges between nodes. Each proposed mapping includes an edge type, a model-generated rationale, and a preliminary confidence score. These proposals are then passed through a human-in-the-loop (HITL) validation pipeline, ensuring that computational efficiency is balanced with cultural authenticity and clinical rigor.

The validation framework is organized into three levels of expert review:
1. **Linguistic validation:** native speakers verify idiomatic usage and contextual appropriateness.
2. **Clinical validation:** mental health practitioners evaluate the diagnostic or therapeutic relevance of the mapping.
3. **Cultural validation:** anthropologists and cultural experts ensure that situated cultural meanings are preserved.

Mappings are presented within a validation interface that encloses confidence scores and provenance, allowing experts to accept, reject, or modify edges. In cases where disagreements arise, structured adjudication rounds are triggered to encourage deliberation and consensus building. When legitimate differences persist, multiple interpretations are retained as parallel edges, thereby avoiding the erasure of cultural diversity.

Through this hybrid approach, computational scalability is combined with expert judgment, resulting in mappings that are broad in coverage and high in quality. However, even after validation, the risk remains that mappings may appear opaque to clinicians or researchers. For the framework to support real-world adoption, every connection must also be interpretable.

### 3.4 Explainability Layer and Transparency Features

To ensure interpretability, CL-PDE integrates explainable AI (XAI) mechanisms that accompany every mapping with layered explanations. Rather than treating edges as opaque links, the system documents why each connection was proposed and how it should be understood in three complementary perspectives: .

- **Linguistic:** highlighting semantic, idiomatic, or metaphorical parallels between expressions.

- **Cultural:** situating expressions within the contexts in which they are commonly used, including regional and social nuances.

- **Clinical:** clarifying how expressions may or may not align with diagnostic categories, and emphasizing when a phrase is non-pathological outside clinical contexts.

For example, when mapping the Hindi expression "*mujhe ghabraahat mehsoos ho rhi hai*", the system surfaces the following: *Linguistic* - a somatic metaphor indexing emotional burden; *Cultural* - commonly used by Hindi speakers for transient stress or sadness; *Clinical* - may correspond to anxiety-related symptoms if persistent, but not diagnostic in isolation.

These explanations are stored alongside provenance metadata and confidence scores so that users can audit each decision. Figure 3 illustrates additional examples of expression–concept mappings with their layered explanations and validation outcomes.

In addition, three transparency mechanisms are implemented to preserve trust and accountability:

- **Confidence scores:** combining model estimates with validator agreement.
- **Provenance tracking:** documenting the origin of each expression (e.g., counseling transcripts, helplines, community data).
- **Alternative interpretations:** retaining multiple valid mappings when consensus is not possible, with clear reasoning provided for each.

Figure 4 shows our explainability interface, which presents clinicians and researchers with multilevel justifications for each mapping. In this way, CL-PDE supports not only accurate and culturally grounded mappings, but also transparent and trustworthy ones that can be meaningfully integrated into clinical and research workflows.

## 4 Methodology

Having outlined the conceptual framework for cross-lingual mental health ontologies, we now describe the methodology for its implementation. The pipeline is organized into three main components. First, data collection and annotation establish a culturally grounded corpus of mental health expressions. Second, graph construction combined with LLM integration aligns these expressions across languages and clinical ontologies. Finally, explainability mechanisms ensure that every mapping remains interpretable and auditable for both clinicians and researchers.

### 4.1 Data Collection and Annotation Protocol

The first step is to construct a corpus that captures the full range of how distress is expressed across Indian languages and contexts. To achieve this, we employ a multi-tier collection strategy:

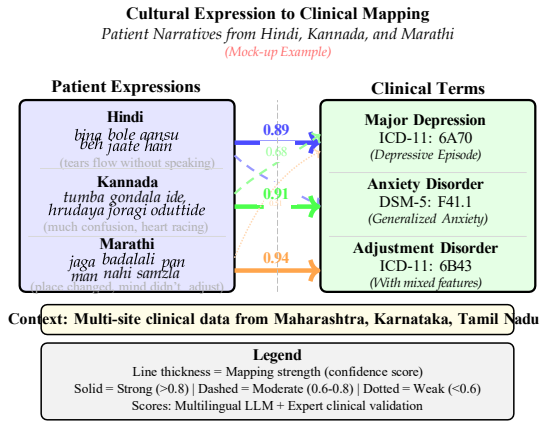

(a) Example mappings from cultural expressions to clinical terminology

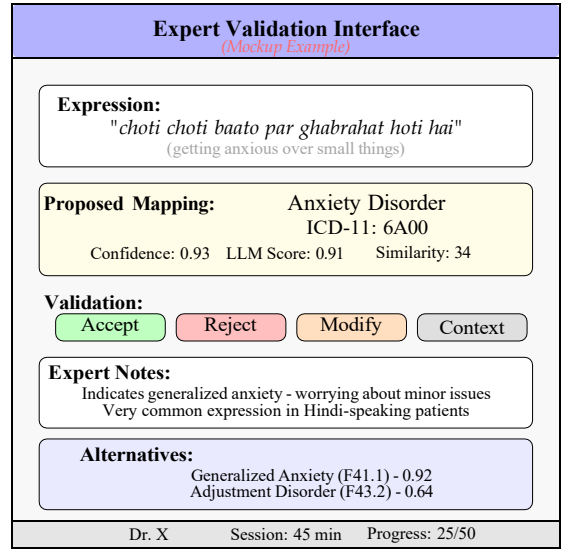

(b) Human-in-the-loop validation interface for culturally grounded panic expression

Figure 3: Examples of cultural expression mapping and validation processes in the CL-PDE framework

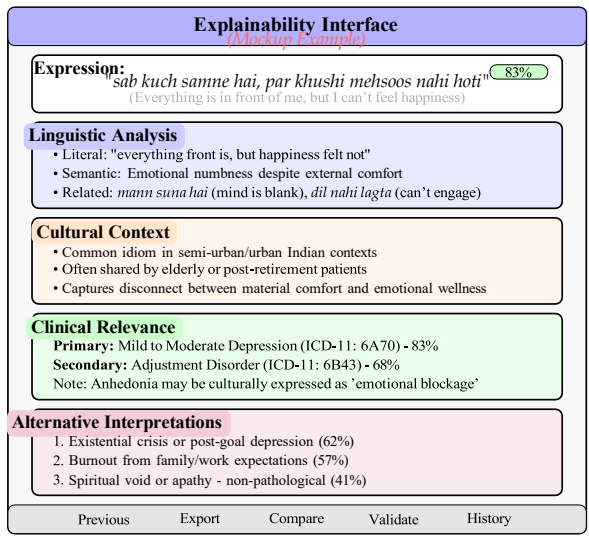

Figure 4: Explainability interface for: "*sab kuch samne hai, par khushi mehsoos nahi hoti*"

1. **Primary sources:** We partner with mental health organizations, counseling centers, and helplines across five states to obtain anonymized transcripts and case narratives. These materials provide direct evidence of how patients describe distress in clinical encounters and cover a wide spectrum of regional and dialectal variation.

2. **Secondary sources:** We supplement clinical data with material from health forums, online support groups, and (with ethical approval) social media discussions. These sources capture colloquial idioms, code-mixed utterances, and emerging metaphors of distress that rarely appear in formal clinical documentation.

3. **Expert consultation:** We conduct structured interviews with clinicians, community health workers, and cultural psychiatrists to document regional idioms and metaphors. Expert input helps connect everyday language with diagnostic categories, while ensuring that culturally specific meanings are retained.

Each expression is then annotated according to a schema designed to balance cross-lingual alignment with cultural specificity. Entries are labeled with semantic categories (e.g., emotion, somatic complaint, behavior), cultural markers (idiomatic or metaphorical usage, references to belief systems), severity indicators (mild vs. severe), and temporal profiles (acute vs. chronic). Confidence scores are recorded to reflect annotator certainty, and disagreements are resolved through multi-annotator discussion. This schema preserves nuance while ensuring interoperability across languages and contexts.

### 4.2 Graph Construction Algorithm

With annotated expressions in place, the next step is to represent them in a graph structure that connects patient language with standardized clinical concepts. Graph construction proceeds as follows:

1. **Extract expression nodes:** Named entity

recognition and phrase-mining techniques identify spans of interest (e.g., idioms, symptoms, metaphors), which are instantiated as expression nodes enriched with their annotation labels.

2. **Generate embeddings and intra-/cross-lingual edges:** Contextual embeddings are computed using multilingual encoders such as mBERT or XLM-R. Similarity measures (cosine distance, alignment models) propose candidate links, which are then filtered and passed to expert validation.

3. **Build expression–concept edges:** For linking expressions to ontology categories (ICD–11, DSM–5, or cultural frameworks), large language models generate candidate mappings along with rationales and uncertainty scores. Human-in-the-loop validation confirms or revises these mappings, ensuring both clinical accuracy and cultural appropriateness.

4. **Enrich with metadata:** All edges are annotated with relation type, provenance, validator confidence, and annotation context. This metadata enables traceability and provides structured input for explainability features.

To maximize efficiency, our HITL system integrates: (i) **active learning**, where uncertain mappings are prioritized for review; (ii) **batch validation**, grouping similar candidates to accelerate expert decisions; (iii) **feedback loops**, updating thresholds based on expert judgments; and (iv) **mismatch resolution**, where structured adjudication ensures consistency across annotators.

### 4.3 Implementation of Explainability

Finally, the ontology is augmented with an explainability layer that makes system decisions transparent. When new expressions are processed, they are aligned to existing nodes using similarity measures or LLM-based semantic alignment. If alignment remains uncertain, provisional nodes are created and annotated. For each edge—whether confirmed or provisional—the system produces a multi-perspective explanation, drawing from both computational signals and annotation metadata.

Explanations are generated through five complementary strategies:

- **Annotation-aware reasoning:** incorporating semantic categories, severity, temporal profile, and cultural markers.

- **Attention visualization:** highlighting words or subphrases most influential in the mapping.

- **Rule-based explanations:** surfacing common idiomatic or metaphorical patterns.

- **Example-based reasoning:** presenting similar validated examples from the corpus.

- **Contrastive explanations:** clarifying why one candidate mapping was chosen over alternatives.

Together, these mechanisms ensure that mappings remain interpretable not only to computational experts but also to clinicians and cultural validators.

### 4.4 Evaluation Plan

Our evaluation spans five dimensions: (i) intrinsic metrics such as graph connectivity, semantic coherence, and inter-annotator agreement (target $\kappa > 0.7$); (ii) extrinsic validation on downstream tasks, including clinical relevance and telepsychiatry deployment; (iii) explainability assessment through measures of user trust and decision transparency; (iv) efficiency of the HITL pipeline; and (v) cultural validity, assessed via expert review and community feedback. This multi-faceted evaluation ensures that the system is not only technically sound but also culturally authentic and clinically meaningful.

## 5 Limitations

Our framework faces several limitations that must be acknowledged. First, the current language coverage focuses on a handful of major Indian languages and may therefore miss the full diversity of regional dialects and tribal languages, as well as the code-mixed expressions that dominate urban digital communication. Explainability also presents challenges, since cultural nuances are often difficult to capture algorithmically, and the quality of explanations can vary depending on available resources across languages. Human validation further poses scalability concerns: expert review is both time-intensive and dependent on the availability of qualified validators who combine cultural knowledge with clinical expertise. The mapping of cultural expressions to clinical terminology is inherently subjective, requiring continuous validation and sometimes yielding legitimate disagreement among experts. Moreover, language itself evolves over time, particularly in digital spaces, demanding regular updates to keep the ontology relevant. Bias remains another concern, as our data sources may overrepresent urban, digitally literate populations despite efforts toward broader representation. Finally, while technical performance provides one measure of success, the ultimate value of these tools will depend on their integration into clinical workflows and their ability to demonstrably improve patient care, a question that requires further validation through clinical trials.

## 6 Conclusion and Future Work

Building cross-lingual mental health ontologies for Indian languages addresses a critical blind spot in healthcare communication. By grounding AI systems in culturally valid representations of distress and providing transparent explanations for all mappings, progress can be made toward inclusive, patient-centric NLP tools that bridge linguistic divides in mental health care. The integration of explainability and human-in-the-loop validation as core components ensures that mappings are not only accurate but also trustworthy and culturally appropriate, which is essential for clinical adoption and patient trust. Looking ahead, the framework will be extended to cover a broader range of Indian languages, including tribal and minority languages, and more sophisticated explanation generation will be developed through large language models fine-tuned on culturally and clinically relevant texts. Interactive explanation interfaces will be designed to allow mappings to be explored at multiple levels of detail, and continuous learning mechanisms will be implemented to improve through ongoing human feedback. Multimodal expressions of distress - such as voice tone and facial expressions - will be incorporated alongside longitudinal tracking to capture how these expressions evolve over time. Culturally-aware dialogue systems will be developed to communicate mental health concepts across language barriers, and rigorous field studies will be conducted to evaluate the framework's impact on clinical workflows, diagnostic accuracy, patient engagement, and therapeutic outcomes.